\documentclass{article}

\PassOptionsToPackage{numbers, compress}{natbib}
\usepackage{tightenitemize}

\usepackage[preprint]{neurips_2024}

\usepackage[utf8]{inputenc} % allow utf-8 input
\usepackage[T1]{fontenc}    % use 8-bit T1 fonts
\usepackage{hyperref}       % hyperlinks
\usepackage{url}            % simple URL typesetting
\usepackage{booktabs}       % professional-quality tables
\usepackage{amsfonts}       % blackboard math symbols
\usepackage{nicefrac}       % compact symbols for 1/2, etc.
\usepackage{microtype}      % microtypography
\usepackage{xcolor}         % colors
\usepackage{graphicx}

\title{Bottom-Up and Top-Down Analysis of Values, Agendas, and Observations in Corpora and LLMs}

\author{
Scott E. Friedman \\
SIFT, USA \\
\footnotesize{\texttt{friedman@sift.net}}
\And 
Noam Benkler \\
SIFT, USA \\
\footnotesize{\texttt{nbenkler@sift.net}}
\And 
Drisana Iverson \\
SIFT, USA \\
\footnotesize{\texttt{dmosaphir@sift.net}}
\And 
Jeffrey Rye \\
SIFT, USA \\
\footnotesize{\texttt{jrye@sift.net}}
\And 
Sonja Schmer-Galunder \\
University of Florida, USA \\
\footnotesize{\texttt{s.schmergalunder@ufl.edu}}
\And
Micah Goldwater \\
University of Sydney, Australia \\
\footnotesize{\texttt{micah.goldwater@sydney.edu.au}}
\And
Matthew McLure \\
SIFT, USA \\
\footnotesize{\texttt{mmclure@sift.net}}
\And 
Ruta Wheelock \\
SIFT, USA \\
\footnotesize{\texttt{rwheelock@sift.net}}
\And
Jeremy Gottlieb \\
SIFT, USA \\
\footnotesize{\texttt{jgottlieb@sift.net}}
\And 
Robert P. Goldman \\
SIFT, USA \\
\footnotesize{\texttt{rpgoldman@sift.net}}
\And 
Christopher Miller \\
SIFT, USA \\
\footnotesize{\texttt{cmiller@sift.net}} \\
}

\newcommand{\sref}[1]{(\S\ref{sec:#1})}

\newcommand{\figref}[1]{Fig.~\ref{fig:#1}}
\newcommand{\secref}[1]{Sec.~\ref{sec:#1}}

\begin{document}

\maketitle
\vspace{-.2in}
\begin{abstract}
\vspace{-.1in}
Large language models (LLMs) generate diverse, situated, persuasive texts from a plurality of potential perspectives, influenced heavily by their prompts and training data.
As part of LLM adoption, we seek to characterize---and ideally, manage---the socio-cultural values that they express, for reasons of safety, accuracy, inclusion, and cultural fidelity.
We present a validated approach to automatically (1) extracting heterogeneous latent value propositions from texts, (2) assessing resonance and conflict of values with texts, and (3) combining these operations to characterize the pluralistic value alignment of human-sourced and LLM-sourced textual data.
\end{abstract}
\vspace{-.2in}
\section{Introduction}
\vspace{-.1in}
As large language models (LLMs) are adopted across healthcare, humanities, and defense sciences, it's increasingly important to measure and manage the values that appear in their outputs.
Measuring values may help us characterize whether model's behavior is consistent with \emph{universalism} (i.e., reflecting a singular or dominant value system), \emph{pluralism} (i.e., attending to a plurality of potentially-conflicting value systems), or something in-between.
This paper presents an approach to analyzing LLMs and datasets to (1) surface a plurality of values from the \textbf{bottom-up} \sref{bottom-up}, (2) measure novel and user-provided values from the \textbf{top-down} \sref{top-down}, and (3) summarize the value dominance and pluralism in a dataset or LLM's output \sref{application}.
Informed by prior work, it's important to characterize language models' value systems at \emph{at least} two levels: (1) the values reflected in [samples of] their training data, and (2) the values reflected in their outputs to diverse prompts and stimuli:

\textbf{1. Values in the data (inputs).}
AI models acquired by large-scale machine learning contain systemic biases rooted in their training corpora \cite{bolukbasi2016man}, and research has shown that these models' biases correlate meaningfully with their training data's temporal origin (e.g., with respect to stereotypes \cite{garg2018word}) and cultural origin (e.g., with respect respect to cultural values \cite{friedman2019relating}).
Consequently, assessing values in training data can help us explain, predict, and manage the values reflected in an LLM's output.

\textbf{2. Values in the generation (outputs).}
LLMs do not ``hold'' values in the same psychological sense that people do, but their outputs may nevertheless consistently support or contradict human values.
Recent work has evaluated the values of LLMs in top-down fashion by (1) using human-based value inventories (e.g., \cite{schwartz2015human}) to directly survey LLMs \cite{miotto2022gpt,johnson2022ghost} or (2) prompting LLMs and then measuring their output's implicit resonance \cite{benkler2023assessing} against cross-cultural value statements \cite{inglehart2000world}.

Most of the above approaches utilize (a) pre-existing, vetted value inventories to directly survey LLMs \cite{schwartz2015human,inglehart2000world,miotto2022gpt,johnson2022ghost} or (b) pre-defined value enumerations to characterize latent values in LLM texts \cite{benkler2023assessing,benkler2024recognizing}.
These \emph{top-down} analyses are useful when we have a \emph{closed set} of relevant values \emph{a priori}.
Unfortunately, top-down analyses are less useful if we are uninterested in surfacing additional, novel values or agendas, or if our \emph{a priori} value enumerations have gaps or blind-spots.
The complementary \emph{bottom-up} value analysis to surface novel values and agendas is a contribution of this work.

We continue with a description of our top-down \sref{top-down} and bottom-up \sref{bottom-up} value analysis strategies and a validation \sref{validation} showing high top-down value analysis accuracy (F1=0.97) and bottom-up value extraction on par with human annotators (\figref{eval}). 
We present applications of combined top-down and bottom-up strategies \sref{application} to demonstrate  efficacy for analyzing value alignment and pluralism.
% \section{Approach}
% \label{sec:approach}

% We describe the NLP approaches to top-down \sref{top-down} and bottom-up \sref{bottom-up} analyses, including the NLP architectures, datasets, and annotation protocols.
\vspace{-.1in}
\section{Approach: Top-Down Analysis of Values \& Themes}
\label{sec:top-down}
\vspace{-0.1in}
Our top-down analyses build on prior pair-wise NLP for stance analysis and value analysis, all based on the broader NLP problem settings of recognizing textual entailment \cite{dagan2022recognizing} and natural language inference \cite{bowman2015large}.
Existing pair-wise approaches of this shape take in a $\langle$\emph{premise}, \emph{hypothesis}$\rangle$ pair and then accurately (and relatively rapidly) predict whether the premise text holds a stance that is $\{$\emph{positive}, \emph{neutral}, \emph{negative}$\}$ with respect to various stance-topic hypotheses \cite{mohammad2016semeval,vast}, or analogously, predict whether the premise text takes a position that is $\{$\emph{resonating}, \emph{neutral}, \emph{contradicting}$\}$ with respect to the various value-hypotheses \cite{benkler2024recognizing,benkler2023assessing}.

We use the $\{$\emph{resonating}, \emph{neutral}, \emph{contradicting}$\}$ analysis of values and themes in this work, and we apply it more broadly: given a novel text $t$, we can assess it against value of interest $v_i$ by running $\langle t, v_i \rangle$, and we can \emph{also} analyze inter-value resonance by running $\langle v_i, v_{i+1} \rangle$ to compute a directed network of resonance and contradiction over a potentially diverse plurality of values.

\textbf{NLP Dataset, Architecture, \& Training:}
To build a combined value- and stance-based dataset, we incorporate training and testing data from the World Value Corpus dataset \cite{benkler2024recognizing} and the VAST dataset \cite{vast}, respectively.
We also utilize our bottom-up theme extraction dataset described below \sref{bottom-up}, including all $\langle$\emph{text}, \emph{theme}$\rangle$ pairs created by the bottom-up human annotators.
The combined training dataset size is 19,804 $\langle$\emph{premise}, \emph{hypothesis}$\rangle$ pairs with varied labels: 7,943 \emph{resonance}, 6,410 \emph{contradiction}, and 5,451 \emph{neutral}.
We trained the DeBERTa-v3 variant \texttt{DeBERTa-v3-base-mnli-fever-anli} on our training data for 4 iterations with learning rate $2e^{-5}$.
\vspace{-.1in}
\section{Approach: Bottom-Up Extraction of Values \& Themes}
\label{sec:bottom-up}
\vspace{-0.1in}
Value inventories \cite{schwartz2015human,inglehart2000world} are useful for psychological and anthropological analyses spanning geography and decades, having been vetted and translated for cross-cultural relevance.
They are also, by nature, incomplete with respect to current events.
For instance, the World Value Survey (WVS) contains agree-disagree prompts regarding reliability of online news sources or trust in the World Health Organization, but it does not---and should not---probe for participants' trust in every social institution, technology, and organization.

Our bottom-up theme extraction uses fine-tuned LLMs (validated in \secref{validation}) to extract three different types of themes.
We exemplify each type from our maternal health value domain \sref{application}.
\vspace{-0.1in}
\begin{enumerate}
    \item \textbf{Observations}: Events or relationships in the world.  These may be true or false, but the text expresses them as facts.  Example: \emph{``Colostrum boosts newborns' immune systems.''}
    \item \textbf{Evaluations}: Judgements on specific topics, including attributions of quality, trust, etiquette, or other dimension of regard.
    Example: \emph{``Hospitals in urban areas are corrupt.''}
    \item \textbf{Agendas}: Statements of ``should [not],'' promoting or justifying principles, norms, and [un]desired world states.
    Example: \emph{``Mothers should feed colostrum to their newborns.''}
\end{enumerate}
We developed these categories after analyzing the WVS: many of its probes are agenda-like, others are evaluation-like, all can be stated as propositions, and probes range from general (e.g., about ``family'') to very specific (e.g., about the WHO).
All three types of themes are in proposition-form, so they could be the subject of agree/disagree survey prompts like the WVS.
This means that the output of bottom-up theme extraction can feed into the existing top-down value assessment (described above), and we can thereby compute a network of bottom-up and top-down themes to help characterize the value alignment and pluralism of a model or corpus.

\textbf{NLP Dataset, Architecture, \& Training:}
No previous dataset exists for this theme/value extraction task, so we developed one for this purpose, using an annotator-in-the-loop approach \cite{schmer2024annotator}.
Three human annotators received annotation guidelines that described the three categories of themes (above) and included 12 fully-worked example paragraphs.
Annotators wrote all of the distinct themes that they encountered in the texts they were assigned, resulting in human annotations for 512 paragraphs from news articles, academic journals, social media, and ethnographic interviews.
Annotators had weekly meetings to collaboratively review difficult examples, using annotator-in-the-loop dataset development practices \cite{schmer2024annotator}.
The dataset was split into (a) a 480-example training set and (b) a 32-example held-out validation set with overlapping annotations from two human annotators.

We fine-tuned two different LLMs---Microsoft's Phi2 and Meta's Llama3 8B-Instruct---for 10 epochs of low-rank adaptation training \cite{hu2021lora}.  The theme-extraction prompt is described in \secref{prompt}.
\vspace{-.1in}
\section{Validation}
\label{sec:validation}
\vspace{-.1in}
% We validated top-down resonance \sref{top-down} and bottom-up extractor \sref{bottom-up} using held-out portions of their respective datasets.

\textbf{Top-down evaluation.}
We ran our fine-tuned DeBERTa-v3 model on the World Value Corpus test set \cite{benkler2024recognizing} and the SemEval 2016 Task 6 Twitter stance detection test sets \cite{mohammad2016semeval}.
For WVC, our model achieved 0.97 micro-F1 (0.92 for resonance, 0.98 for neutral, and 0.97 for contradiction), tied with the state-of-the-art RoBERTa model \cite{benkler2024recognizing}.
For SemEval 2016, our model achieved 0.78 micro-F1 for Task A (0.8 for resonance, 0.68 for neutral, 0.8 for contradiction) and 0.71 micro-F1 for Task B (0.65 for resonance, 0.78 for neutral, and 0.65 for contradiction), out-performing the SemEval 2016 competition winners for both Task A (F1=0.67) and Task B (F1=0.56).
This is evidence that our top-down value-resonance strategy \sref{top-down} produces high-quality predictions.

\begin{figure}[t]
\centerline{\includegraphics[width=\textwidth]{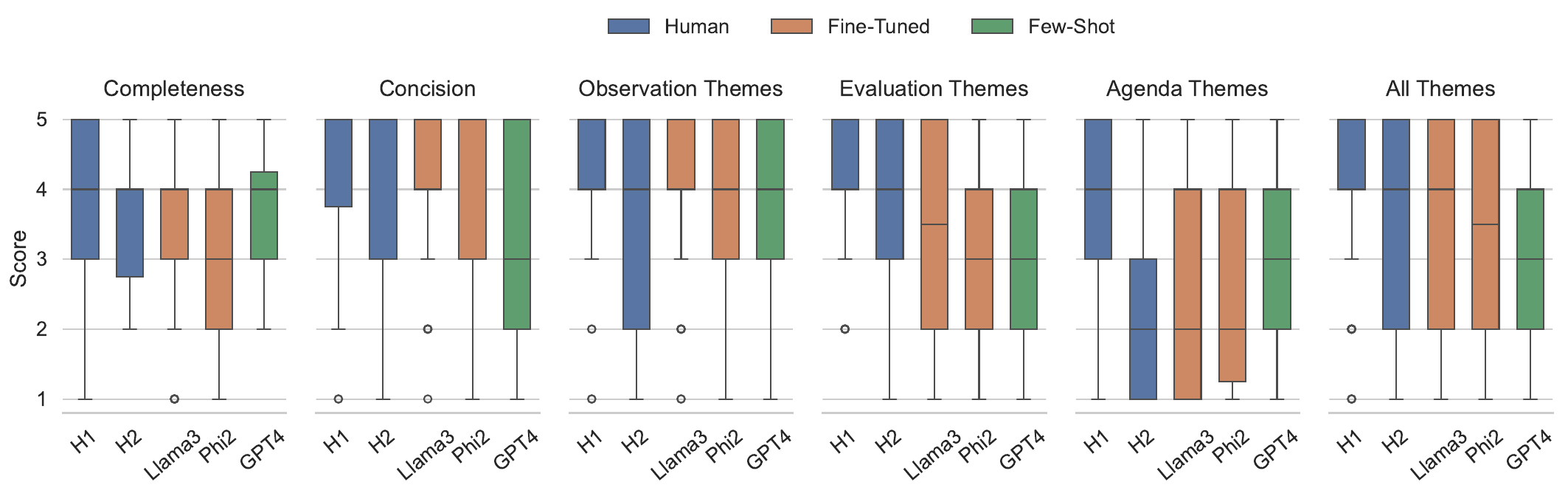}}
\centering
\caption{Evaluation results of human judges rating human and machine theme extraction.}
\label{fig:eval}
\vspace{-.2in}
\end{figure}

\textbf{Bottom-up evaluation.}
Two human judges who were not involved with the bottom-up annotation process \sref{bottom-up} received the annotation guidelines and used a web interface to make quality judgments on the themes extracted by humans and machines from a held-out test set.
The two judges were blinded with respect to who or what extracted the themes (i.e., whether it was a human or a machine).
The judges rated sets of extracted themes on a scale of 1 (poor) to 5 (excellent) along dimensions of \emph{completeness} (how well the results captured all the themes in the text) and \emph{concision} (how well the results minimize unnecessary content).
In addition to evaluating the full extractions, they also rated each \emph{individual} theme extracted on 1-to-5 quality scales.
Results are shown in \figref{eval}, comparing quality judgments across dimensions of the work of the two human raters (\textbf{H1} and \textbf{H2}), our two fine-tuned models (\textbf{Llama3} and \textbf{Phi2}), and 12-example few-shot results from \textbf{GPT4}.
The only statistically significant results are that (1) \textbf{GPT4} performed significantly worse on concision and (2) \textbf{H1} produced higher-quality agendas and total themes compared to other humans and machines.
Finally, the human judges predicted whether a human or a machine extracted the themes, and their accuracy at human-machine prediction was no better than chance (F1=0.52).
These combined results suggest that our bottom-up theme and value strategy \sref{top-down} produces high-quality extractions.

\vspace{-.1in}
\section{Example Applications}
\label{sec:application}
\begin{figure}[t]
    \centering
    \begin{minipage}{0.5\textwidth}
        \centering
        \includegraphics[width=\textwidth]{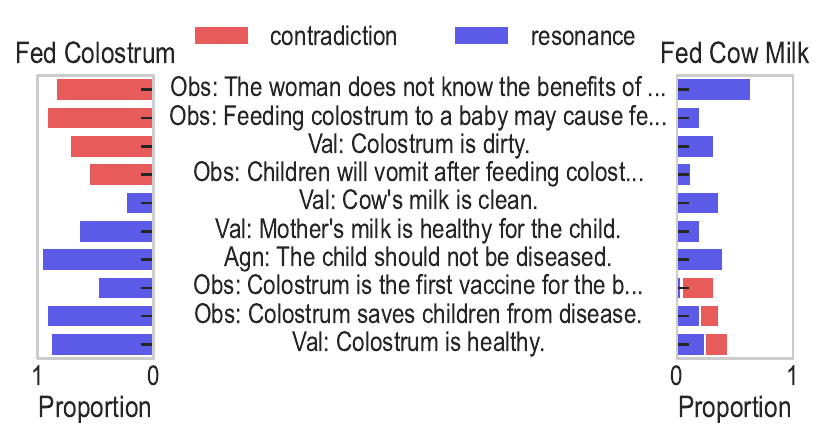} % first 
        \vspace{-0.3in} 
        \caption{Analysis of RISE colostrum interviews.}
        \label{fig:rise}
        \vspace{0.2in} 
        \centering
        \includegraphics[width=\textwidth]{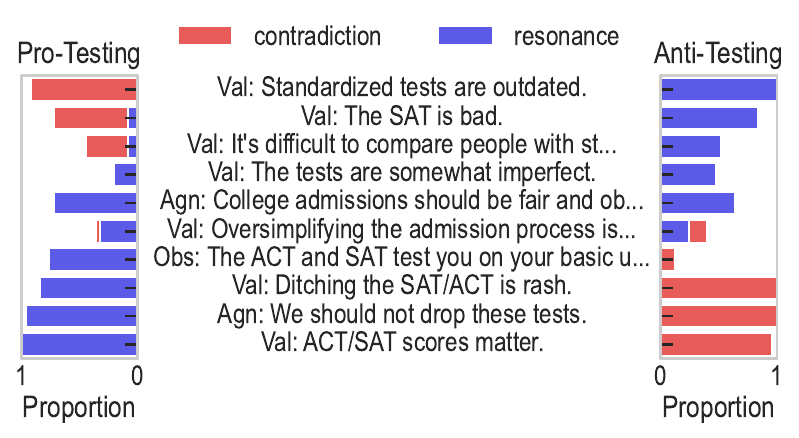}
        \vspace{-0.3in} 
        \caption{Analysis of GPT4 on standardized tests.}
        \label{fig:testing}
    \end{minipage}\hfill
    \begin{minipage}{0.5\textwidth}
        \centering
        \includegraphics[width=\textwidth]{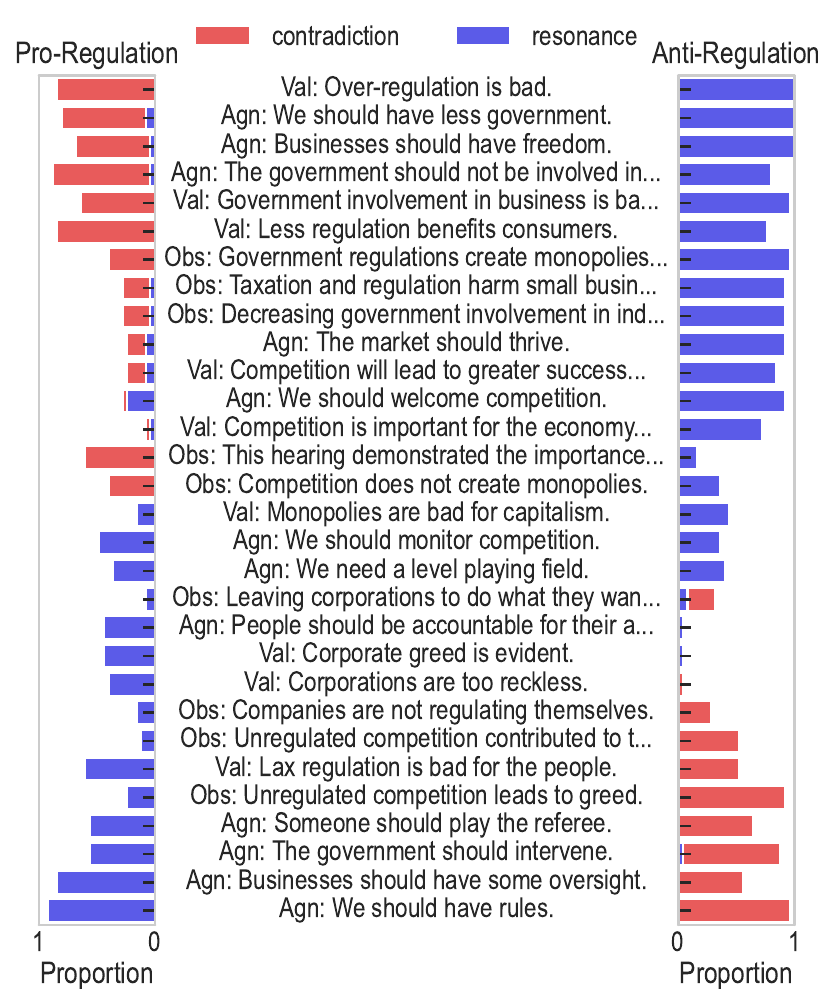}
        \vspace{-0.3in} 
        \caption{Analysis of GPT4 on gov oversight.}
        \label{fig:regulation}
    \end{minipage}
    \vspace{-.2in}
\end{figure}

\vspace{-.1in}
We combine top-down \sref{top-down} and bottom-up \sref{bottom-up} analyses of themes and pluralism, on two levels analysis: (1) analyzing datasets from ethnographic interviews with human participants and (2) analyzing argumentative LLM outputs focused at a specific topic.
Each of our datasets includes \emph{``pro''} and \emph{``anti''} positions for a given topic, so we perform a comparative analysis of the plurality (or universality) of machine-extracted values for each position.
For each analysis, we perform (1) bottom-up extraction of themes from texts and then (2) top-down characterization of how themes relate (i.e., resonance, contradiction, neutrality) to both \emph{``pro''} and \emph{``anti''} positions.
All results are shown in Figs. \ref{fig:rise}, \ref{fig:testing}, and \ref{fig:regulation}, listing the most relevant (i.e., non-\emph{neutral}) observations (``Obs''), evaluations (``Val''), and agendas (``Agn'') that were extracted by our system.
For each theme, we plot the proportion of texts (between 0 and 1.0) that resonate and contradict the theme, indicating whether a position is divided, neutral, or consistent with respect to that theme.

\textbf{Values in Maternal and Child Health Interviews.}
Our dataset analysis utilizes a subset of the RISE corpus \cite{hashmi2023integrating} where researchers interviewed mothers in India about why a woman might choose to feed her newborn colostrum\footnote{Colostrum is breast milk expressed after birth, highly-concentrated with nutrients but often thick and yellow.} (25 documents) \emph{vs.} cow-milk (25 documents).
\figref{rise} shows the results: the most resonant observations and evaluations for feeding cow-milk (at right) are that women might not know the benefits of colostrum, colostrum may cause fever or vomiting, and colostrum is ``dirty.''
Each of these themes are contradictory to the pro-colostrum position (at left).
Both positions support that cow's milk is ``clean'' and that mother's milk is healthy, but differ on the health of colostrum.

\textbf{Values in GPT4 Comments about Standardized Testing.}
We prompted GPT4 to write 50 comments for a fictitious news article ``Colleges Continue to Drop SAT/ACT Requirements'' \sref{gen-prompt} adopting the positions of (1) pro- and (2) anti-standardized-tests.
\figref{testing} shows results, where pro- and anti-positions disagree on evaluations of the relevance and quality of SAT/ACT, but both strongly agree on an agenda for admission fairness, and moderately agree that the tests are imperfect.

\textbf{Values in GPT4 Comments about Government Oversight of Business.}
We prompted GPT4 to write 50 comments for a fictitious news article ``Partisan Conflict Surges Following Hearing on Corporate Taxation and Regulation'' \sref{gen-prompt} adopting the positions of (1) pro- and (2) anti-government regulation of business.
\figref{regulation} shows results, including positional disagreement about the value and consumer impact of rules, freedom, greed, and regulation.
Both positions support agendas for fairness and monitoring competition, and both evaluate monopolies as harmful.

\vspace{-.2in}
\section{Conclusion}
\label{sec:conclusion}
\vspace{-.1in}
This paper presented a novel approach to bottom-up value discovery from text \sref{bottom-up} that achieves human-level value extraction \sref{validation} and a state-of-the-art top-down assessment of value resonance \sref{top-down}.
We apply our approach on three domains---one human dataset and two LLM-generated corpora---to perform fully-automated value analyses.
These analyses confirmed the values we expected (i.e., when we prompted the LLM to take a stance) and they also surfaced additional, unexpected values that were expressed in the texts.
This permits us to analyze LLMs' value alignment and pluralism with respect to (1) \emph{a priori} value hypotheses from human operators and (2) latent values from the text, in support of comparative read-outs.
For future work, we plan to apply this approach in diverse domains and at larger scales, and to help characterize how the plurality of values expressed in LLM training data (or in prompts) impact the values expressed in LLMs.

\begin{ack}
The research was supported by funding from the Defense Advanced Research Projects Agency (DARPA HABITUS W911NF-21-C-0007-04). The views, opinions and/or findings expressed are those of the authors and should not be interpreted as representing the official views or policies of the Department of Defense or the U.S. Government.
\end{ack}

\bibliographystyle{ieeetr}
\bibliography{values}

\appendix

\section{Appendix / supplemental material}

\subsection{Ethical Considerations}

The values extracted from texts in our approach are not necessarily those intended by the human or machine that authored the text.
Consequently, values extracted or predicted by the machine to resonate with an author's texts should not be attributed to the authors; rather, they should be regarded as plausible value implications.
This approach should not be applied to make moral judgments, ascribe ethics to authors, or classify individuals based on their socio-cultural commitments, since this is outside the bounds of our validation.

\subsection{Limitations}

Our models operate on unstructured texts alone, and not imagery, audio, video, or structured data.
Further, our models are trained on texts of paragraph-size, so they do not have the larger context of a full news article, journal article, or fictional work when making extractive or associative value judgments.
It also does not have access to an overview of current events, so texts concerning topics that change rapidly---such as geopolitical events and pop culture---may be inaccurately or incompletely characterized by our approach.
Our model has been validated on a hold-out set of 32 examples that spans multiple domains, but we plan to validate it on additional domains and input formats in future work.

\subsection{Bottom-Up Theme Extraction Prompt for Fine-Tuned LLMs}
\label{sec:prompt}

We used the following prompt for bottom-up theme extraction with LLMs, populating the \texttt{input\_text}.
To encode categorized themes, we used a newline-delimited sequence of themes, associating each \texttt{theme\_text} with its \texttt{theme\_category} (Observation, Evaluation, or Agenda) and the \texttt{attribution} of the theme (i.e., whether it's held by the author or by another entity mentioned in the text.
During training, we specified the themes in this fashion, and during prediction/validation, we parse the LLM-generated themes from this format into structured data.

{\small
\begin{verbatim}
Instruct: List themes from the text from the perspective of the author and
others.  For each theme, label the type (Observation, Evaluation, or Agenda)
and the perspective in parentheses.  No duplicates.

Input: <input_text>
Output:
<theme_text_1> (<theme_category_1> by <attribution_1>)
...
<theme_text_n> (<theme_category_n> by <attribution_n>)
\end{verbatim}
}

\subsection{Document Generation Prompt for GPT-4}
\label{sec:gen-prompt}

We used the following prompt for document generation with GPT-4, populating \texttt{source\_article}, \texttt{target\_evaluation}, and \texttt{target\_agenda} with the values in table \ref{tab:generation-vals}. 

{\small
\begin{verbatim}
Write me five varied comments in response to an online news article with the 
headline '<article>' that (1) <agenda> and (2) <evaluation>.
Make these as casual as possible.
\end{verbatim}
}

\begin{table}[h!]
\centering
\scriptsize
\caption{Generation Prompt Inputs}
\label{tab:generation-vals}
\begin{tabular}{p{2cm}|p{2cm}||p{1cm}|p{3.4cm}|p{3.4cm}}
% \hline
\textbf{Topic} & \textbf{Article} & \textbf{Stance} & \textbf{Agenda} & \textbf{Evaluation} \\ \hline\hline
Market Regulation  & Partisan Conflict \newline Surges Following  \newline Hearing on Corpo- & Pro & Government involvement in business and industry should be increased & Unregulated competition is harmful \\
& rate Taxation and \newline Regulation & Anti & Government involvement in business and industry should be decreased & Competition is beneficial \\ 
&&&& \\
Standardized\newline Testing & Colleges Continue to Drop SAT/ACT Requirements & Pro & Standardized tests should be required for college and university admission decisions & Standardized tests are reliable for normalizing and predicting academic success \\
& & Anti & Standardized tests should not be required for college and university admission decisions & Standardized tests are unreliable for normalizing and predicting academic success \\
\end{tabular}
\end{table}

Each unique prompt was run 5 times for a total of 25 comments per unique prompt, using the following model settings:
\begin{description}
    \item[model:] gpt-4 
    \item[temperature:] 1.0
    \item[max\_tokens:] None
    \item[timeout:] None
    \item[max\_retries:] 2
\end{description}

\end{document}